\documentclass[11pt,onecolumn]{article}

\usepackage{amsmath}
\usepackage{amsfonts}
\usepackage{amssymb}

\usepackage[ansinew]{inputenc}
\usepackage{graphicx,color}
\usepackage{url}
\usepackage{latexsym,enumerate}
\usepackage{booktabs}
\usepackage{framed}
\usepackage{subfigure}
\usepackage{multirow} 
\usepackage{color}
\usepackage{colortbl}
\definecolor{gray1}{rgb}{0.8,0.8,0.8}
\definecolor{gray2}{rgb}{0.95,0.95,0.95}
\newcommand{\RR}{\mathbb{R}}

\newcommand{\hbit}{\hspace*{1.5pt}}

\begin{document}

\title{The Shortlist Method for Fast Computation of the Earth Mover's Distance
and Finding Optimal Solutions to Transportation Problems}%
\date{}%
\author{Carsten Gottschlich\thanks{C. Gottschlich and D. Schuhmacher are with 
the Institute for Mathematical Stochastics, University of G\"ottingen, 
Goldschmidtstr. 7, D-37077 G\"ottingen, Germany. 
C.~Gottschlich is also with the Felix-Bernstein-Institute for Mathematical Statistics in the Biosciences,
University of G\"ottingen, Goldschmidtstr. 7, D-37077 G\"ottingen, Germany.
Email: {gottschlich@math.uni-goettingen.de} and {dominic.schuhmacher@mathematik.uni-goettingen.de}}
\and Dominic Schuhmacher\footnotemark[1]}%

\maketitle

\begin{abstract}
Finding solutions to the classical transportation problem is of great importance,
since this optimization problem arises in many engineering and computer science applications.
Especially the Earth Mover's Distance is used in a plethora of applications 
ranging from content-based image retrieval, shape matching, fingerprint recognition, object tracking
and phishing web page detection to computing color differences in linguistics and biology.
Our starting point is the well-known revised simplex algorithm, 
which iteratively improves a feasible solution to optimality. 
\emph{The Shortlist Method} that we propose substantially reduces the number of candidates inspected 
for improving the solution, while at the same time balancing the number of pivots required.
Tests on simulated benchmarks demonstrate a considerable reduction 
in computation time for the new method as compared to the usual revised simplex algorithm 
implemented with state-of-the-art initialization and pivot strategies.
As a consequence, the Shortlist Method facilitates the computation of large scale transportation problems 
in viable time. In addition we describe a novel method for finding an initial feasible solution 
which we coin \emph{Modified Russell's Method}.
\end{abstract}

\section*{Keywords}
Transportation problem, revised simplex method, earth mover's distance, EMD, 
Wasserstein distance, Monge-Kantorovich problem

\section{Introduction}

Finding solutions to the classical transportation problem is of great importance,
since this optimization problem arises in various guises in many real world and theoretical situations.
They occur as subproblems in larger problems, e.g.
the warehouse location problem or the traveling salesperson problem
and also in a variety of engineering and computer science applications, 
such as content based image retrieval \cite{RubnerTomasiGuibas2000},
automatic scene analysis \cite{RicciZenSebeMesselodi2013}
or for the discrimination between real and artificial fingerprints \cite{GottschlichHuckemann2014}.
A more extensive discussion of such applications is given in Section \ref{secApplications}.

The problem was first described by Monge in 1781 \cite{Monge1781} in somewhat different form 
and has been analyzed by many researches including Kantorovich, Hitchcock, Koopmans
and especially Dantzig \cite{Dantzig1963,Dantzig1990}, the father of the simplex algorithm.    
The solution of this problem is the fundamental ingredient for computing
the Earth Mover's Distance \cite{RubnerTomasiGuibas2000} in computer science and
the Wasserstein distance, also known as Mallows or Kantorovich distance
in statistics and physics, see Chapter~6 in \cite{Villani2008}.

In order to give a quick and intuitive description of the various facets of the transportation problem 
and the revised simplex algorithm we often use an economic interpretation, 
which of course will not reduce the scope of the described algorithms and their applications in any way.
The problem can be summarized as follows.

Consider a consortium of $m$ production and $n$ consumption facilities of a certain good. 
For simplicity these are also referred to as origins and destinations. 
Suppose that there is a certain supply of $a_i>0$ available at origin $i$, and 
there is a certain demand of $b_j$ at destination $j$. 
The cost for transporting a unit of the good from $i$ to $j$ shall be given by arbitrary $c_{ij} \in \RR$. 
Borrowing the illustration from Chapter~3 in \cite{Villani2008}, the production facilities might be 
Parisian bakeries cooperating with caf{\'{e}}s (consumption facilities), where the good transported are baguettes, 
and the cost incurred is the actual transportation cost.  
It is assumed that total supply equals total demand, i.e. $\sum_{i=1}^{m} a_i = \sum_{j=1}^{n} b_j$.
The objective is then to determine a transportation plan $X = (x_{ij})_{1 \leq i \leq m, 1 \leq j \leq n}$ such that 
all producers and consumers are satisfied and that the total cost is minimized. In other words
\begin{eqnarray}
\textrm{\bf{minimize  }} \sum_{i=1}^{m} \sum_{j=1}^{n} c_{ij} x_{ij}& &\\
\textrm{\bf{subject to  }} \sum_{j=1}^{n} x_{ij} = a_i& & \textrm{for } i = 1, ..., m,\\
\sum_{i=1}^{m} x_{ij} = b_j& & \textrm{for } j = 1, ..., n,\\
x_{ij} \ge 0& & \textrm{for all }i, j.
\end{eqnarray}

A dual formulation can be obtained as follows. 
Suppose that a carrying company offers to take over the good 
from the consortium for a price of $u_i \in \RR$ per unit at origin $i$ and 
to hand it back at destination $j$ for a price of $v_j \in \RR$ (any prices may be negative). 
In order for the carrier to be competitive, it needs to set prices $u_i, v_j$ 
so that $u_i + v_j \leq c_{ij}$ for all $1 \leq i \leq m$, $1 \leq j \leq n$. 
Following \cite{LuenbergerYe2008} we refer to the difference $r_{ij} = c_{ij} - u_i - v_j$ 
as \emph{relative cost} incurred when the consortium takes over 
the transportation from $i$ to $j$ itself rather than commissioning the carrier. 
The carrier would like to maximize its profit $\sum_{i=1}^{m} a_i u_i + \sum_{j=1}^{n} b_j v_j$ subject to 
the price constraint. Standard duality theory, e.g.\ Chapter~4 in \cite{LuenbergerYe2008} relates 
the solutions of the two problems to one another (provided one of them exists) 
and shows that the optimal values of the objective functions are the same.

The rest of the paper is organized as follows.
In Section~\ref{secTransportationAlgorithm} we first give a non-technical description of the revised simplex algorithm for solving the transportation problem; for a more detailed presentation see \cite{LuenbergerYe2008}. Then we discuss crucial aspects in various subsections, starting with pivot strategies (Subsection~\ref{ssecPivot}), and passing from cycle finding to treating initialization methods (Subsections~\ref{ssecCycFinding} and~\ref{ssecInitMethods}, respectively).
In Section \ref{secShortlist}, we introduce the new Shortlist Method 
for solving the transportation problem.
Benchmark tests reported in Section \ref{secResults} clearly show the advantage of the proposed method
over the existing ones. 
We conclude with a discussion of the results in Section \ref{secConclusions}
and review relevant application scenarios in Section \ref{secApplications}.

\section{The Transportation Algorithm} \label{secTransportationAlgorithm}

Using the simplex approach the transportation algorithm consists of two stages: 
first, an initial transportation plan $X$ is constructed such that Equations~(2--4) are satisfied.
Second, the initial plan is iteratively improved until the optimal solution is obtained.

At any time the current feasible plan consists of $m+n-1$ ``active'' origin/destination 
pairs $(i,j)$ between which a positive amount $x_{ij}$ is transported (in a degenerate case 
there might be pairs with zero amount, but we exclude this case in our description). 
We will refer to them as basis pairs or basis entries.

For each iteration in the second stage a basis entry is replaced by a ``better one''. 
For this we first compute the ``dual'' prices $u_i$ and $v_j$. 
In the context of the simplex method, these are also known as simplex multipliers. 
Starting with an arbitrary value, e.g.\ setting $u_1 = 0$, all other prices are determined 
by solving the equations $u_i + v_j = c_{ij}$, where $(i,j)$ are basis entries. 
A property well-known as basis triangularity sees to it that every origin 
and every destination gets a price assigned in this way.

A new basis entry is then selected as a so-called pivot element by finding a non-basis pair $(i,j)$ 
that has negative relative cost $r_{ij} = c_{ij}-u_i-v_j$, meaning 
that the consortium can transport goods more cheaply from $i$ to $j$ 
by itself than by commissioning the carrier. 

Next, a cycle of changes starting in $(i,j)$ is determined by alternately scanning rows and columns 
for basis entries until a cycle is complete, which again is bound to happen by basis triangularity. 
Assuming that all amounts $x_{i'j'}$ at basis entries are positive (the non-degenerate case), 
there is a maximal positive amount $\theta$ which we can alternately add and subtract 
from the values $x_{i'j'}$ when following the cycle, starting with addition 
for the first value $x_{ij}$. Since the cycle alternates between following rows and columns,
the procedure preserves Equations (2--3).

After this, one of the $x_{i'j'}$ has been reduced to $0$ 
and we remove the corresponding pair $(i',j')$ from the basis 
(if several values have been reduced to zero, we remove the first such entry, 
but are then dealing with a degenerate case).
The basis still has exactly $n + m - 1$ entries, 
and we proceed with the next iteration, continuing until there are no entries 
with negative relative cost any more. In this case we have reached an optimum.

\subsection{Pivot Strategies} \label{ssecPivot}

When selecting a pivot element to enter the basis, all non-basis entries with relative cost $r_{ij} < 0$ are candidates. According to Dantzig's criterion, the most negative one is chosen.
To the best of our knowledge it is an open question whether a better criterion 
for selecting one of these candidates can be formulated 
in order to minimize the number of pivot operations until optimality is reached.

If the algorithm is applied to solve real-world transportation problems,
the goal of a practical implementation is typically to minimize the runtime on a computer.
Our analysis has shown that \emph{two} key factors determine the runtime: the number of pivot operations 
and the number of elements for which relative costs are computed in order to select pivot elements.

The former can be made small by computing the relative costs for all non-basis entries
which in turn maximizes the latter (`matrix most negative' strategy).
The other extreme is to perform the pivot operation immediately 
after discovering the first candidate (`first negative' strategy). 
In this way, the second factor is minimized at the cost of an increase of the first.
A more balanced strategy is to compute the relative costs for all non-basis entries
of a row and then choose the most negative among these candidates (`modified row most negative strategy')
or go on with the subquent row, if no candidate has been discovered.
In the next iteration of the algorithm, continue with the first row not considered in the previous one.
The latter strategy outperformed the others in our tests,
which corroborates earlier findings reported 
by \cite{SrinivasanThompson1973} and by \cite{GloverKarneyKlingmanNapier1974}.

\subsection{Finding Cycles} \label{ssecCycFinding}

\begin{figure}[!th]
 \begin{center}
 \begin{tabular}{cc}
\vtop{%
\vskip-1ex
\hbox{%
\includegraphics[width=0.4\textwidth]{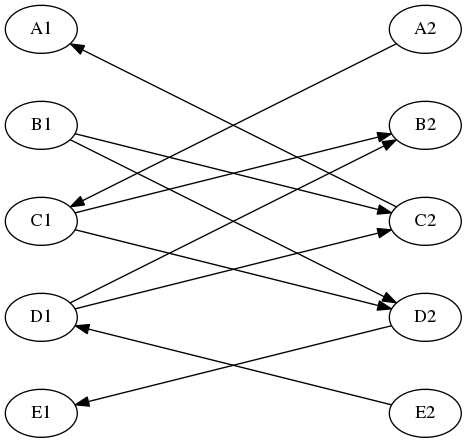}%
}%
} &
\vtop{%
\vskip-1ex
\hbox{%
\includegraphics[width=0.4\textwidth]{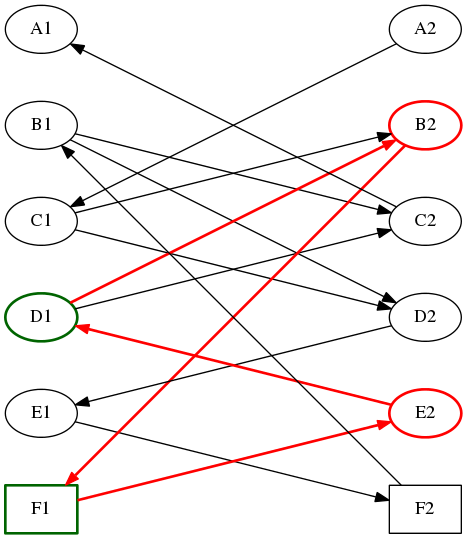}%
}%
} \\
& \\
$X =
\left( \begin{array}{ccc}
  & A &  \\
B & C & D \\
  &  & E
\end{array} \right)$ &
$X =
\left( \begin{array}{ccc}
  & A &  \\
B & C & D \\
\mathbf{F} &  & E
\end{array} \right)$ \\
\end{tabular}
 \end{center}
 \caption{\label{fig:dfs} Each graph corresponds to the transport plan shown below.
          Directed edges are drawn with arrows. 
          The direction from left to right indicates a 'same row' relation between basis entries,
          right to left shows a 'same column' relation. 
          The graph on the left becomes cyclic by adding the pivot element F to the basis (right).
          \label{figGraph}}
\end{figure}

The procedure of finding cycles of changes can be translated 
into a depth-first search (DFS) \cite{Sedgewick2003} 
on the following directed graph (see also Figure~\ref{fig:dfs}):
Each basis entry corresponds to two vertices: one vertex with the basis entries
in the same row as incoming edges and the basis entries in the same column as outgoing edges,
and a second vertex with the basis entries in the same column as incoming edges and the
basis entries in the same row as outgoing edges. The graph is weakly connected, acyclic and bipartite.
By adding the (two copies of the) pivot element, the graph becomes cyclic 
and DFS is an efficient method for discovering the (up to mirroring) unique cycle.
Since each basis entry is connected to all other basis entries
in the same row and the same column, no other data structure 
is needed to store the graph than a list of basis entries for each row and for each column.

Considering the example shown in Figure ~\ref{fig:dfs}, 
we begin with the transport plan and graph on the left.  
Next, we insert F as pivot element (right) and discover the cycle starting in F1 with depth-first search.
Along the cycle, the minimum of all nodes on the right side of the graph determines the amount of change $\theta$
which is substracted from B and E (red) and added to D and F (green). 
One of the two elements B or E will leave the basis.
F2 was not required during the pivot operation, 
but alternatively it would have been possible to use the complementary cycle F2 $\to$ B1 $\to$ D2 $\to$ E1 $\to$ F2 
instead, leading to the same result. 
Basis elements A and C remain unchanged during this pivot operation.

\subsection{Initialization Methods} \label{ssecInitMethods}

In the subsequent comparison of methods for constructing an initial feasible solution 
(stage one in the transportation algorithm), 
we take the following established procedures into account.
If a method generates fewer than $m+n-1$ basis entries (degenerate case), 
we complement them by adding the right number of entries $(i,j)$ in such a way that all basis entries are connected, 
i.e.\ there are other basis entries in the same row or the same column, but no cycles are formed 
and their values $x_{ij}$ remain zero.

\textbf{Northwest Corner Rule.} Suppose we list all origins from $i=1$ to $i=m$ as rows 
and all destinations from $j=1$ to $j=n$ as columns. 
This rather naive rule
starts in the top left corner and allocates the maximum possible amount to $x_{11}$, 
i.e. the minimum of $a_1$ and $b_1$. If there remains supply at origin $1$, 
we move to the right and assign to $x_{12}$ maximum possible amount. 
Otherwise if the demand at destination $1$ was larger than the supply, 
we move one cell down and continue with assignment $x_{21}$. 
And in case that $a_1$ is equal to $b_1$, we move directly to $x_{22}$.
In this way, we iterate over all origins and destinations,
and we obtain a solution satisfying Equations (2-4).

\textbf{Least Cost Rule or Matrix Minimum Rule.} This simple rule 
determines in each iteration 
the minimum cost entry $c_{ij}$ among all origins with remaining supply 
and among all destinations with remaining demand, and assigns 
the maximum possible amount to $x_{ij}$ until all requirements are met.

\textbf{Houthakker's Method of Mutually Preferred Flows.} 
The idea of Houthakker's mutually preferred flows \cite{Houthakker1955} 
is somewhat similar to the least cost rule. 
For all origins that have any supply left, the minimum cost $c_{ij}$ of the corresponding row is determined,
and likewise for all destinations that have any demand left, 
the minimum cost $c_{ij}$ of the corresponding column is detected.
If an entry $(i, j)$ is both row and column mimimum, 
the maximum feasible amount is assigned to $x_{ij}$.
A difference to the least cost method is that more than one entry can enter the basis 
in each iteration. 

\textbf{Vogel's Approximation Method.} The basic idea of Vogel's approximation method \cite{ReinfeldVogel1958} 
is to compute the opportunity costs: 
for each not yet exhausted origin and for each remaining destination,
take the difference between its smallest cost and its second smallest cost.
This idea is also the key ingredient for computing bids and raising prices 
in the auction algorithm \cite{BertsekasCastanon1989}.
In each iteration of Vogel's approximation method, the row or column 
with the maximum opportunity cost is selected 
and for the minimum $c_{ij}$ in that row or column, 
the maximum possible value $x_{ij}$ is allocated.

\textbf{Russell's Method.} Russell \cite{Russell1969} proposed an approach to approximate Dantzig's criterion. 
In each iteration denote by $I$ the set of origins $i$ that have any supply left 
and by $J$ the set of destinations $j$ that have any demand left. 
Then determine $w_i = \max_{j \in J} \hbit c_{ij}$ for every $i \in I$ 
and $y_j = \max_{i \in I} \hbit c_{ij}$ for every $j \in J$. 
The quantities $w_i$ and $y_j$ are supposed to approximate 
the simplex multipliers $u_i$ and $v_j$ (see Section~\ref{secTransportationAlgorithm}).
Using these estimates, Russell computes in each 
iteration $(i,j) = \mathrm{arg\,min}_{i \in I, \hbit j \in J} \, (c_{ij} - w_i - y_j)$ 
and allocates the maximum possible amount to $x_{ij}$.

\textbf{Modified Russell Method.} In this paper, we propose a modification of Russell's method 
which outperforms the original version on our benchmarks: 
instead of updating $w_i$ and $y_j$, we compute these values once at the start.
Next, we compute a cost matrix $D$ with $d_{ij} := c_{ij} - w_i - y_j$
and then, we apply the least cost rule to this matrix $D$.
The proposed modification saves a lot of computational time in each iteration
by not updating $w_i$ and $y_j$ and performs much better in comparison 
to the orginal Russell method (see Figure \ref{figRuntime}).

\textbf{Weighted Frequency Method.} Eight years before Russell, 
Habr \cite{Habr1961} proposed a related method which he callled weighted frequency method. 
Let $mr_i$ be the mean cost of row $i$ and $mc_j$ the mean cost of column~$j$. 
According to Habr's method, we define a matrix $F$ with cost entries $f_{ij} := c_{ij} - mr_i - mc_j$.
The transportation plan is established by choosing $x_{ij}$ in each iteration pursuing
the matrix minimum rule applied to $F$ and assigning the maximum possible amount to $x_{ij}$.
Habr provides a nice theoretical justification for his method: suppose for each possible 
entry $(i,j)$ we consider each possible combination $(r,s)$ with $r \neq i$ and $s \neq j$.
The question whether it is beneficial to include $x_{ij}$ in the transportation plan
is answered by comparing the costs $c_{ij} + c_{rs}$ with the costs $c_{is} + c_{rj}$
for all combinations $(r,s)$. Habr showed that summing up the 
differences $c_{ij} + c_{rs} - c_{is} - c_{rj}$ over all possible combinations
is equivalent (up to a constant) to computing the matrix $F$.

\textbf{Row Minimum Rule and Modified Row Minimum Rule.}
These two rules \cite{GloverKarneyKlingmanNapier1974} iterate over the rows (origins)
and determine for each row $i$ the column (destination) with positive unassigned demand $b_j$ 
which has the minimum transportation cost $c_{ij}$.
The difference between both rules is that modified row minimum rule assigns at most one entry $x_{ij}$ 
per row and then resumes with the next row. 
The row minimum rule in contrast repeatedly determines the minimum for row $i$ 
until the supply of origin $i$ is completely distributed and only then it continues with the next row.

\textbf{Column Minimum Rule and Modified Column Minimum Rule.} 
These two rules work exactly as the two previous described methods with rows and columns exchanged.

\textbf{Alternating Row Column Minimum Rule.}
This initialization method combines the modified row minimum rule 
and the modified column minimum rule by alternating between rows and colums.

\textbf{Two Smallest in Row Rule.} 
The two smallest in row rule \cite{SrinivasanThompson1973} can be regarded 
as a variant of the modified row minimum rule that assigns two instead of one entries per row and iteration.

\section{The Shortlist Method} \label{secShortlist}

\begin{figure}
  \begin{center}
    \includegraphics[width=0.48\textwidth]{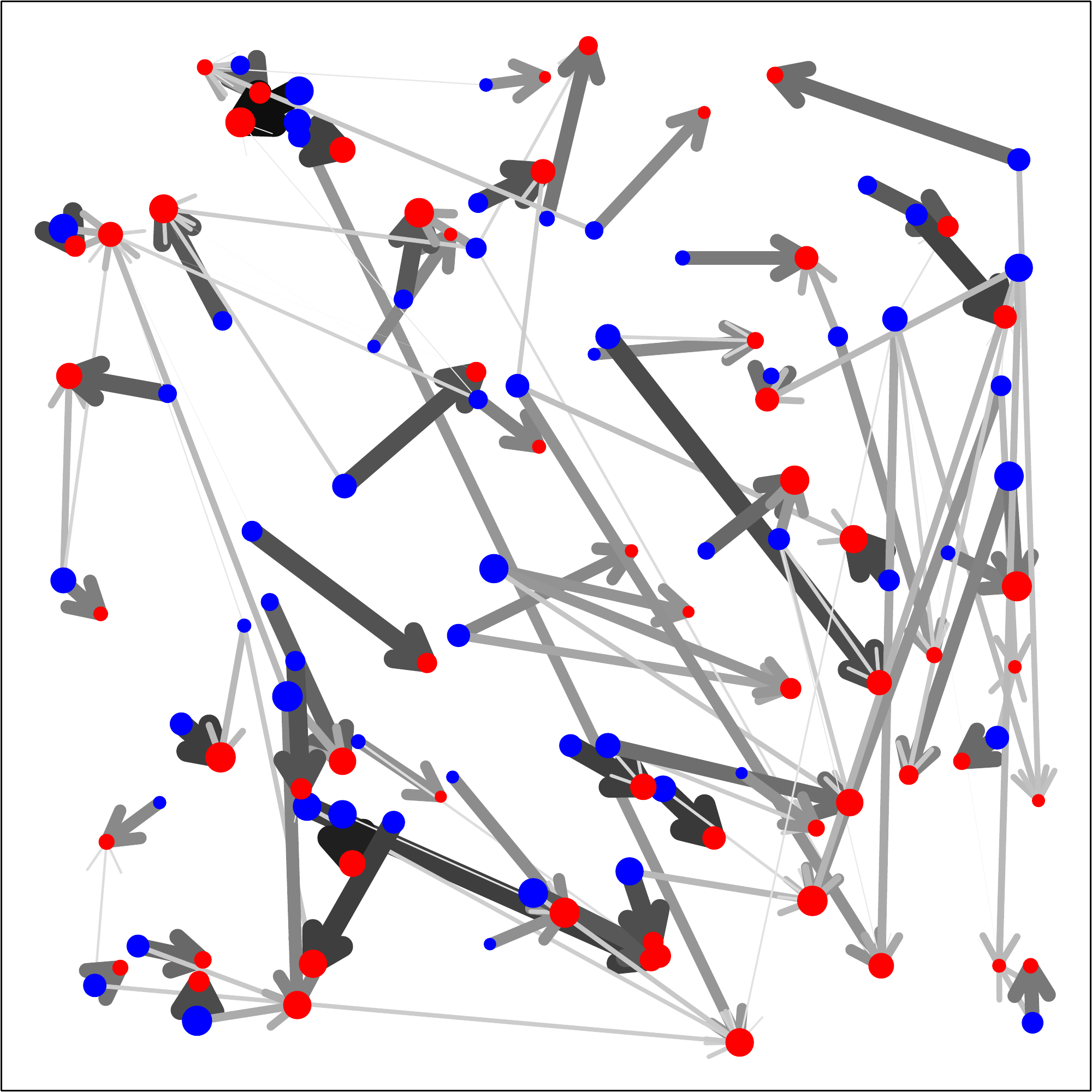}
    \includegraphics[width=0.48\textwidth]{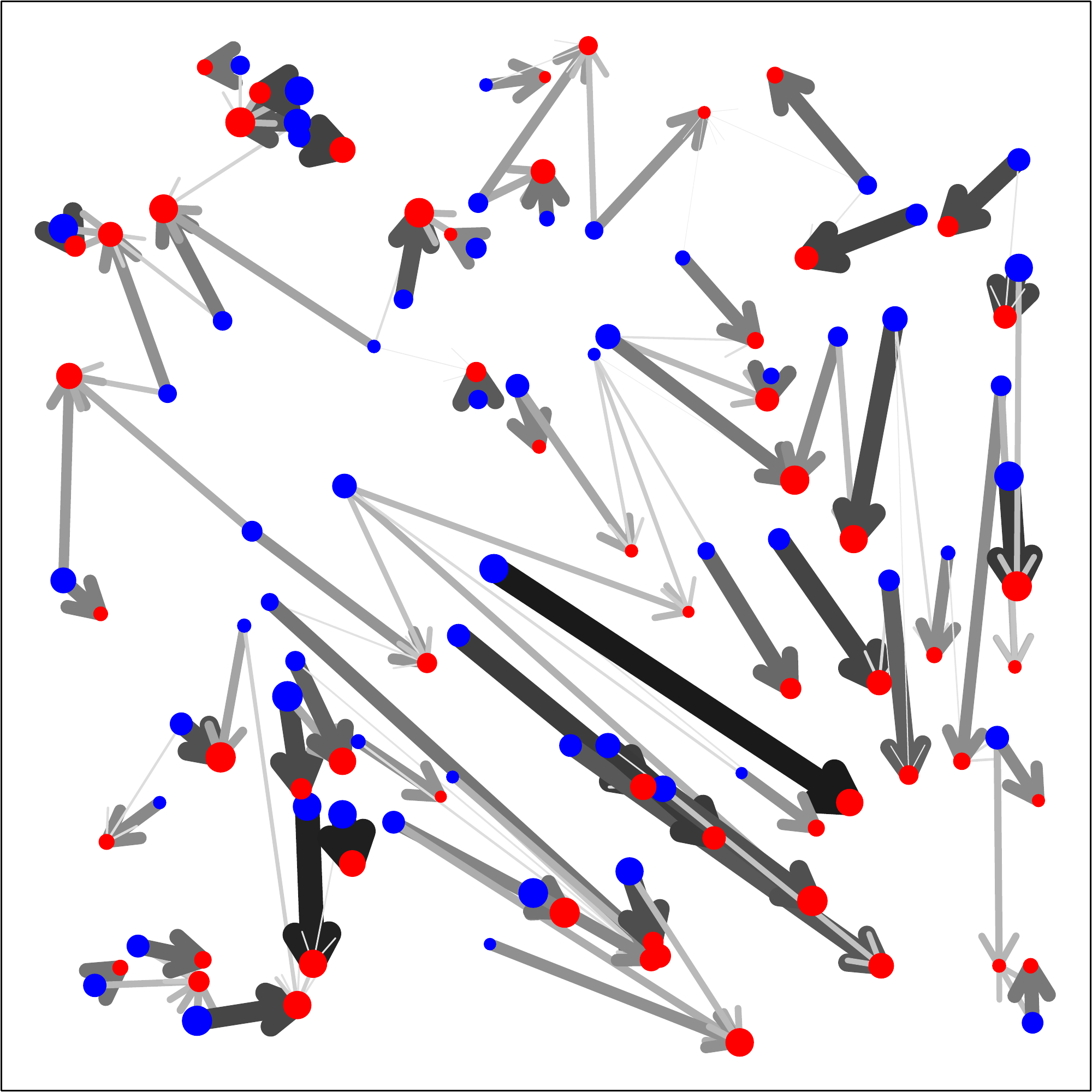}
  \end{center}
  \caption{Visualization of a transportation plan for a very small example with 60 origins (blue)
           and 60 destinations (red) at the end of 
           phase two (left; initial feasible solution derived from shortlists) 
           and at the end of phase four (right; global optimal solution).
           The diameters of the circles correspond to the mass at these origins and destinations.
           A greater width and darker color of arrows indicates a larger amount 
           of mass being transported.
           \label{figFlow}}
\end{figure}

As described in the previous section, the simplex-based transportation algorithm consists of two stages: 
an initialization phase to find a feasible solution and a convergence phase
in which the current solution is iteratively improved to optimality.
The \textit{Shortlist Method} introduces an additional phase in between these two.
The main steps of the Shortlist Method can be outlined as follows:
\begin{enumerate}
\item A shortlist is created for each origin containing only a small fraction 
of all possible destinations.
\item An initial feasible transportation plan is derived from these shortlists 
      (for an example see Figure~\ref{figFlow}, left).
\item The transportation plan is improved towards optimality based on the shortlists.
\item The transportation plan is improved to global optimality based on the complete matrix
      (for an example see Figure~\ref{figFlow}, right).
\end{enumerate}
The crucial part is the third step in which the shortlist search for a new basis entry balances 
the computional burden between the number of elements for which relative costs are calculated
and the number of pivot operations performed.

More precisely the Shortlist Method uses as parameters the length $s$ of the shortlists 
and two decision criteria $k$ and $p$. The four steps are carried out as follows.

At the beginning, for each origin $i$, a list of $s$ destinations 
with the lowest transportation costs is created, containing the index $j$ of the destination 
and the corresponding costs. 
This shortlist is sorted ascendingly according to costs by QuickSort \cite{Sedgewick2003}.

Next, we iterate over all not yet exhausted origins $i$ 
and assign the maximum feasible amount to $x_{ij}$ with the smallest costs $c_{ij}$
among all destinations $j$ in the shortlist of $i$. If no such destination is available any more, 
the minimum over the remaining $j$ is chosen. The latter is usually only necessary for very small shortlist lengths.

In the third phase, we improve the transportation plan $X$ iteratively 
considering batches of consecutive shortlists. 
Starting from the first shortlist not considered in the previous iteration, 
we compute relative costs $r_{ij} = c_{ij} - u_i - v_j$ 
for non-basis entries until $k$ candidate entries with negative $r_{ij}$ have been discovered 
or $p$ percent of all shortlists have been searched. Then the batch ends. 
We choose the entry with the most negative relative cost for performing a pivot operation, 
i.e.\ we add the entry to the basis, compute a cycle of changes and remove another entry 
from the basis as detailed in Section~\ref{secTransportationAlgorithm}. 
Then we go the next iteration. Whenever the last shortlist has been used, 
we continue by reusing the first one. 
If at any point no more candidates are discovered, phase three is terminated.

In the final phase, complete rows are searched instead of shortlists
and if a row contains at least one candidate, the most negative one is chosen; 
i.e.\ we perform the simplex-based transportation algorithm as described 
in Section~\ref{secTransportationAlgorithm} with the `modified row most negative' pivot strategy 
until the optimum is reached.

\section{Simulation Results} \label{secResults}

In order to evaluate the performance of the described initialization methods 
as a function of the number of origins and destinations,
a benchmark was generated in the following way: 
On an empty grid of size $512 \times 512$, the $x$- and $y$-coordinates of locations for $n$ origins
and $n$ destinations were chosen independently and uniformly at random while avoiding double allocations.
Amounts $a_i$ and $b_j$ were chosen independently and uniformly at random between 0 and 255. 
A final adjustment step ensures the equality of the sum over all $a_i$ and the sum over all $b_j$.
The cost matrix $C$ contains as entry $c_{ij}$ the Euclidean distance between origin $i$ and destination $j$.
100 examples are generated for each number $n$ of origins and destinations from 100 to 3000 in steps of 100.

We make the generated transportation problem examples available for download, 
so that other researchers can reproduce the results and test other methods on this benchmark. 
The URL for the download is: \url{www.stochastik.math.uni-goettingen.de/TransportBenchmark}

\begin{figure}[!th]
  \begin{center}
    \includegraphics[width=0.9\textwidth]{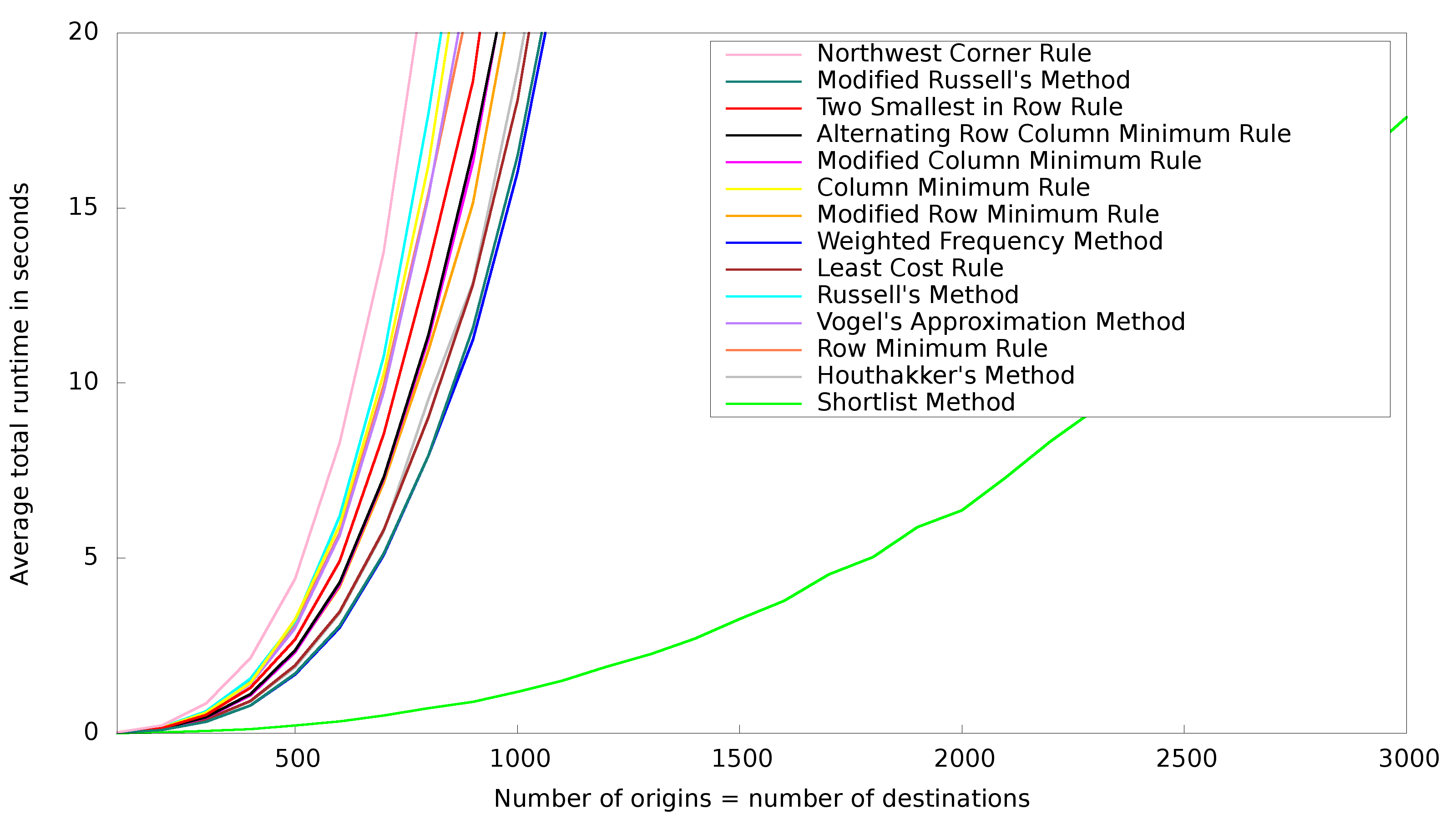} \\
    \includegraphics[width=0.9\textwidth]{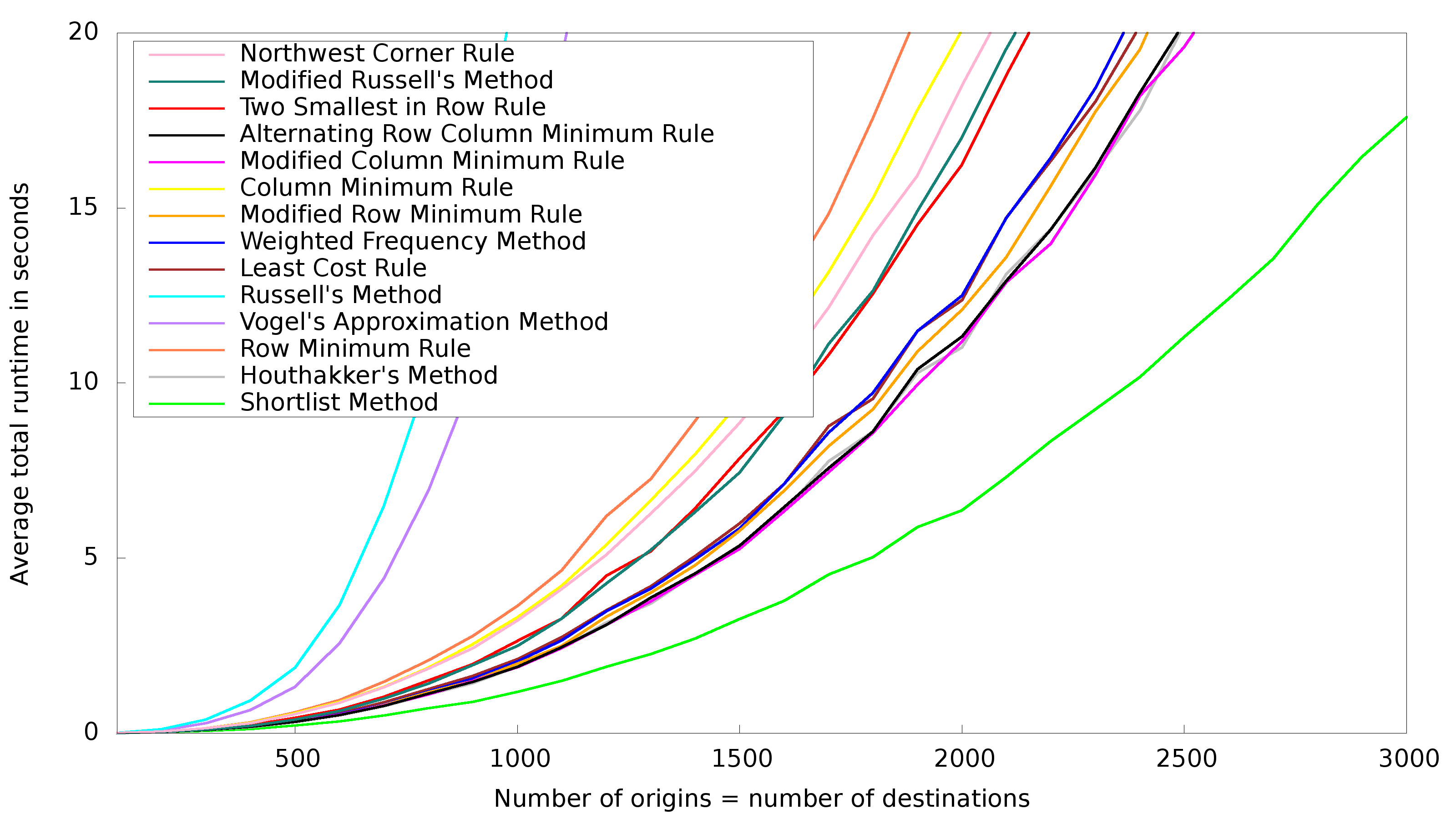}
  \end{center}
  \caption{Comparison of the Shortlist Method to other methods. 
           Depicted are total runtimes in seconds 
	   (for each method and each number of origins averaged over 100 solved transportation problems)
	   for various initialization methods from the literature 
           combined with one of two pivot strategies: 
           matrix most negative (top) and modified row most negative (bottom).
	   The total runtime encompasses the runtime for finding an initial basis 
           and the runtime for the simplex iterations.
           \label{figRuntime}}
\end{figure}

All initialization methods were implemented to the best of our knowledge and 
optimal solutions were computed using the same revised simplex implementation for all methods.
In Figure~\ref{figRuntime}, we report total runtimes including 
the runtime for finding an initial basis and the runtime for the simplex iterations.
The total runtimes are averaged over the 100 examples for each~$n$.
The implementations are written in Java and were tested using one core of an Intel Core i7 CPU with 3.20GHz.

We observe that the Shortlist Method outperforms the other methods by a rather large margin. While for other initialization methods it is clearly preferable to use the ``modified row most negative pivot strategy'' (compare the remark in Subsection~\ref{ssecPivot}), this makes hardly a difference for the Shortlist Method. We may attribute this to the fact that this choice of the pivot only enters in step 4 of the Shortlist Method. However, by the end of step 3 the solution is already so close to optimality that step 4 does not have much influence on the total computation time.

The aforementioned parameters of the Shortlist Method were chosen in the following way. 
An additional set of examples was created with 30 examples for each $n$. 
For each parameter, a set of a few possible choices were defined, 
and in total about thirty of their combinations were used 
for computing initial bases and optimal solutions on this training set. 
In this way, we obtained the following rough rule of thumb: 
\vspace*{3mm}

\begin{center}
\parbox{0.9\textwidth}{\textit{Shortlist length:} $s := 15$ for $n \leq 200$, 
then an increase of $s$ by another $15$ for each doubling of $n$. 
More precisely, $s := 15 + \hbit \lfloor 15 \cdot \log_2 (n/200) \rfloor$ for $n > 200$.}
\vspace*{3mm}

\noindent
\parbox{0.9\textwidth}{\textit{Stop criteria:} 
(i) $k := s$ candidates.
(ii) $p := 5\%$ of shortlists are searched at most in one iteration. 
}%
\end{center}
\vspace*{3mm}

Although these parameter values have been trained, we consider them to be rather ad hoc, 
as they were chosen informally and by considering a few choices only. 
We understand this as a proof of concept of the Shortlist Method 
and as a first step towards determining good universal parameters that only depend on the problem size. 
There are clearly situations, where one has the opportunity to train the method 
to more specific features of the problem at hand, e.g.\ when comparing images to a larger database. 
Then we expect our method to perform even considerably better than suggested by the above simulations. 

Table \ref{tabCompEmdistLpSolve} gives a comparison of our implementation of the shortlist method with two other programs: 
the original C code by Rubner used via the R package emdist \cite{emdist} 
and lp\_solve~\cite{lpsolve} by Berkelaar and others,
a general purpose mixed integer linear programming solver
(which accounts to some extent for its long runtime).

Last but not least, let us note that we have also compared the different approaches 
on various collections of real and randomly generated images, 
and the respective performances were largely confirmed.

\begin{table}[t]
\begin{center}
 \begin{tabular}{|c|r @{.} l|r @{.} l|c|} \hline
 Problem size & \multicolumn{5}{c|}{Method} \\ \hline
 &\multicolumn{2}{c|}{lp\_solve \cite{lpsolve}}&\multicolumn{2}{c|}{emdist \cite{emdist}}&Shortlist \\ \hline 
$100\times100$&     0&1360 &  0&0616 & 0.0054 \\ \hline
$200\times200$&     1&1839 &  0&1507 & 0.0246 \\ \hline
$300\times300$&     4&3854 &  0&5705 & 0.0634 \\ \hline
$400\times400$&    10&6491 &  1&8974 & 0.1245 \\ \hline
$500\times500$&    22&4806 &  4&8668 & 0.2254 \\ \hline
$600\times600$&    40&4955 &  9&0441 & 0.3525 \\ \hline
$700\times700$&    67&5250 & 17&0948 & 0.5269 \\ \hline
$800\times800$&   104&1458 & 28&5478 & 0.7411 \\ \hline
$900\times900$&   145&6244 & 42&1987 & 0.9436 \\ \hline
$1000\times1000$& 203&5568 & 62&3756 & 1.2314 \\ \hline
\end{tabular}
\caption{Comparison of the shortlist method with {lp\_solve}~\cite{lpsolve} and emdist~\cite{emdist}.
         Runtimes in seconds averaged over 100 solved transportation problems.
         \label{tabCompEmdistLpSolve}}
\end{center}
\end{table}

\section{Discussion} \label{secConclusions}

In this paper, we have presented a novel method for solving the classical transportation problem
in its full generality (with an arbitrary cost matrix) 
based on the simplex algorithm 
and we have compared the proposed Shortlist Method to best methods known in the literature.
The results for various problem sizes demonstrate the potential of the novel approach,
which clearly outperforms all the other methods on the considered benchmark 
and indicates that the performance difference between the Shortlist Method and
the second best alternative grows with increasing problem size. 

To substantiate this conjecture we have simulated additional sets of 10 examples 
for each of the six best performing methods in the lower panel of Figure~\ref{figRuntime} 
in combination with each of the problem sizes $n = 400$, $800$, $1600$, $3200$, $6400$, and $12800$. 
Based on the literature we have expected polynomial growth of the time complexity of the problem 
with an exponent that is somewhat below $3$, i.e.\ a runtime of roughly the form $r = c \hbit n^q$ 
for some $q \in [2,3]$. Since this implies that $\log(r) = \log(c) + q \log(n)$ one can expect 
a roughly linear relation, when drawing the logartihm of the runtime as a function of the log problem size. 

\begin{figure}[!th]
  \begin{center}
    \includegraphics[width=0.65\textwidth]{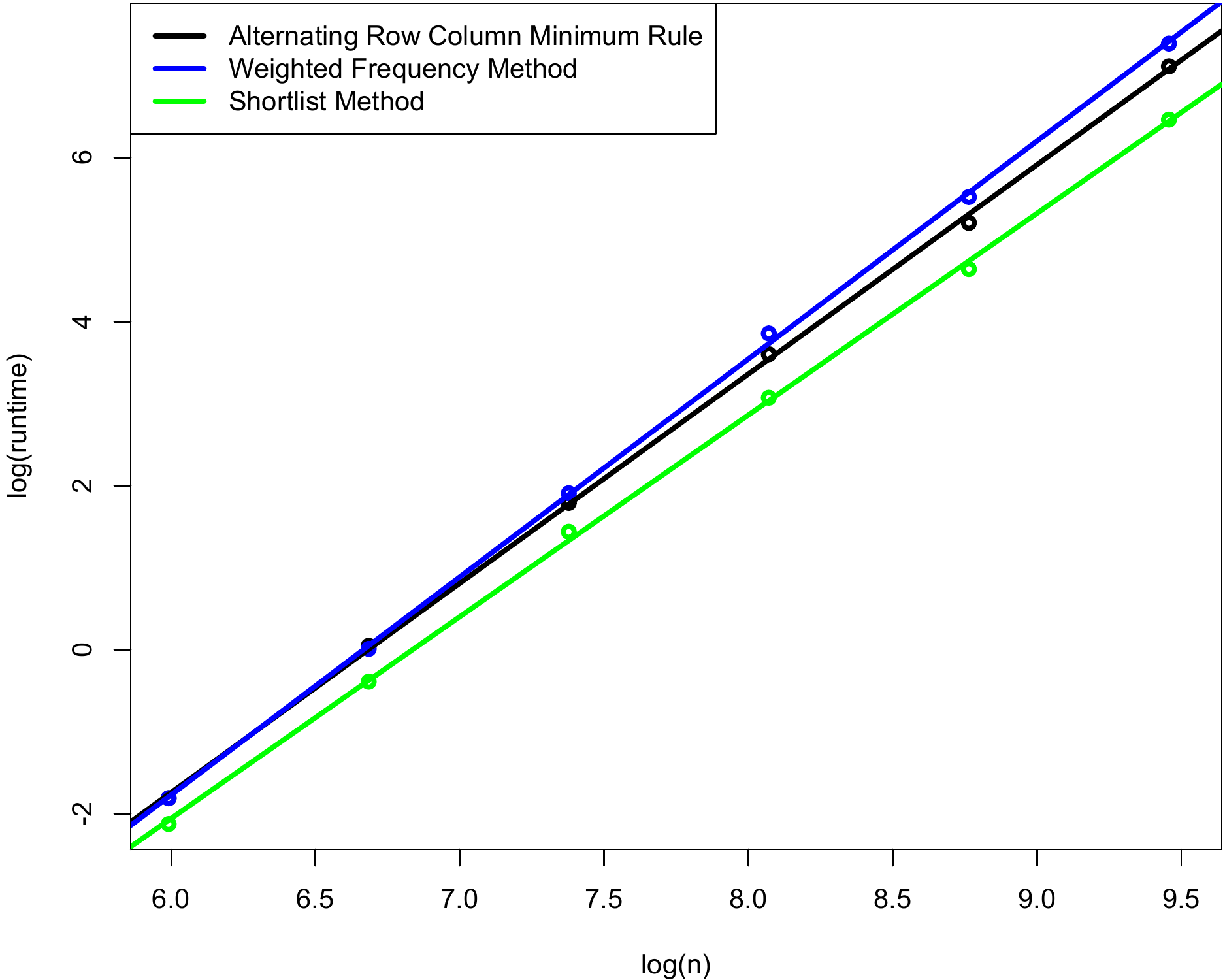}
  \end{center}
  \caption{Comparison of the Shortlist Method to two main competitors. 
           Depicted as circles are the logarithms of total runtimes in seconds 
	   (for each method and each number of origins averaged over 10 solved transportation problems)
           depending on the logarithm of the problem size (circles). The lines have been fitted
           by least-squares regression.
           \label{figloglog}}
\end{figure}

As we can see from Figure~\ref{figloglog} this idea works out quite well. 
We only plot the results for the Shortlist Method and two competing methods as the other four competitors 
would overlap large parts of the two that are given. The circles indicate the results from our simulations, 
whereas the lines have been fitted by least-squares regression. 
Note that the lines fit the simulation data very well. 
The slopes of the lines provide estimates for the exponents $q$. 
These are given numerically in Table~\ref{tabexponents} for all seven methods, 
together with p-values for testing whether the slope is different from the $q$ obtained for the Shortlist Method. 
The p-values are based on statistical tests for comparing slopes in an ANCOVA model, see \cite[Chapter 13]{Faraway2004}. 
Since they are so small, it seems highly likely 
that the Shortlist Method has in fact a better time complexity than the other methods.

\begin{table}[!htb]
\setlength{\tabcolsep}{7pt}
\renewcommand{\arraystretch}{1.3}
\begin{center}
\scalebox{0.85}[0.95]{
\begin{tabular}{|l||l|l|l|l|}
  \hline
  Method                              & \scalebox{0.8}[1]{\hspace*{-2.5pt}factor [$\cdot 10^{-8}$]} & \scalebox{0.8}[1]{\hspace*{-2.5pt}\textbf{exponent}} & p-value  & \scalebox{0.8}[1]{\hspace*{-2.5pt}signif.}    \\
  \hline
 Shortlist Method                    & $5.0026$ & \bf 2.4591 & ---  &     \\
 Alternating Row Col. Minimum Rule   & $3.9526$ & \bf 2.5510 & 0.009090 & $**$   \\
 Modified Column Minimum Rule        & $3.4778$ & \bf 2.5667 & 0.002624 & $**$   \\
 Modified Row Minimum Rule           & $3.1825$ & \bf 2.5915 & 0.000312 & $***$  \\ 
 Least Cost Rule                     & $2.3954$ & \bf 2.6362 & 0.000005 & $***$ \\
 Weighted Frequency Method           & $2.0282$ & \bf 2.6574 & $< 10^{-6}$ & $***$ \\
 Houthakker's Method                 & $1.7119$ & \bf 2.6594 & $< 10^{-6}$ & $***$ \\
  \hline
\end{tabular}
}
\end{center}
\setlength{\tabcolsep}{6pt}
\renewcommand{\arraystretch}{1}
\vspace*{4mm}

\caption{We assume a relation of $r = c \hbit n^q$ between computation time $r$ and problem size $n$. 
Shown are estimates of the factor $c$ (to be multiplied by $10^{-8}$) and the exponent $q$, 
together with p-values for the comparison of exponents for the Shortlist versus other methods. 
Significance levels correspond to the usual classification: $0 < *** < 0.001 < ** < 0.01 < * < 0.05 < \cdot < 0.1$.
         \label{tabexponents}}
\end{table}

Let us also compare the performances of the best six competitors for our original benchmark 
to earlier performance studies from the literature. 
Based on the results considered in the lower panel of Figure~\ref{figRuntime}, i.e.\ based on 
problem sizes up to 3000, several initialization methods performed similarly: 
the modified column minimum rule, Houthakker's method and the 
alternating row column minimum rule, 
followed by the modified row minimum rule,
the least cost rule and the weighted frequency method.

These results confirm earlier findings reported 
in \cite{SrinivasanThompson1973} and in \cite{GloverKarneyKlingmanNapier1974} on other benchmarks,
with one exception: the least cost rule (also known as matrix mimimum rule) 
performed among the best competitors in our test and finished among the slowest methods
in \cite{SrinivasanThompson1973}. A possible explanation is our implementation 
which sorts all matrix entries once ascendingly by transportation costs and then iterates 
over the list until the initial solution is obtained. 
This procedure is more efficient than determining the matrix cost minimum 
in each iteration by scanning all remaining origins and destinations.
Analogously, we can explain the advantage of the proposed modified Russell's method 
over the original Russell's method. The speedup gained by the avoidance of scanning large parts 
of the complete matrix in each iteration clearly outweighs a possible quality loss 
of the initial solution by not updating the quantities $w_i$ and $y_j$ which are supposed to approximate 
the simplex multipliers $u_i$ and $v_j$ (see Section~\ref{secTransportationAlgorithm}).

Further research includes a systematic large-scale simulation study to determine 
good universal parameter settings depending only on easy-to-determine features of the problem such as problem size. 
Also we would like to investigate to what extent computation times and orders of complexity 
can be improved when comparing images within a homogeneous database, 
where one has the possibility to train the parameters to the expected type of transportation problem.

In either case we believe that there is still much room for improvement of the results obtained above. 
We expect these findings to prepare the ground for applications in pattern recognition, 
computer vision and image processing,
where solving the transportation problem has so far been considered as intractable
due to the problem size and the runtimes of existing methods when applied to (smaller) raw gray scale images
or feature descriptors like curved Gabor filter bank responses \cite{Gottschlich2012}. 
A selection of further applications is contained in the next section.

\section{Applications of the Transportation Problem} \label{secApplications}

Solving transportation problems efficiently is of great importance in many different fields of application.
We would like to give an idea of the relevance of fast algorithms
by discussing a selection of specific examples.

\paragraph{Detection of Phishing Web Pages}

The earth mover's distance (EMD) has been applied 
for the detection of phishing web pages by Fu et \textit{al.} \cite{FuWenyinDeng2006}.
Screenshots are taken from banking websites and potential phishing sites
and the visual similarity is measured using the EMD.
If an antiphishing system automatically compares thousands or millions of websites,
the speed of each comparison is an important factor and can become the bottleneck of the system.
In this application scenario, the speedup by the shortlist method can make a huge difference. 
E.g.\ if web sites are compared at a resolution of $100 \times 100$ pixels,
this correponds to a problem with an approximate dimension of 5000 origins and 5000 destinations.

\paragraph{Linguistics}

The EMD has been applied as a measure of dissimilarity
when comparing the distribution of color names among 110 different languages 
\cite{VejdemojohanssonVejdemoEk2014}.
Notably, computation of EMDs for 2300 language vectors took the authors about one week 
using an industrial strength LP solver \cite{CPLEX}. 
Due to the computational complexity, they
refrained from evaluating the 23,982 speaker response vectors.

\paragraph{Content-based Image Retrieval}

Since the early days of retrieving images from large databases,
the EMD has been applied for comparing histograms and signatures \cite{RubnerTomasiGuibas2000}.
Pele and Werman proposed a thresholded ground distance which is an EMD variant \cite{PeleWerman2009}.
For content-based image retrieval, thresholding the ground distance 
has a positive effect on the retrieval accuracy \cite{LvCharikarLi2004}.

\paragraph{Fingerprint Recognition}

In the area of fingerprint recognition, the EMD has been applied for
discriminating between real and synthetic fingerprint images 
based on minutiae histograms \cite{GottschlichHuckemann2014}.
These 2-dimensional minutiae histograms capture the minutiae distribution
as a fixed-length feature vector which is invariant to rotation, translation
and the variations in the number of minutiae. 
Scale invariance can be achieved by scaling input fingerprint images
or minutiae templates to the size of adult fingerprints at a fixed resolution, e.g. 500 DPI.
Fingerprints of adolescents can be enlarged using an age-dependent scaling factor 
as described in \cite{GottschlichHotzLorenzBernhardtHantschelMunk2011}.

\paragraph{Performance Evaluation of Multi-Object Filters}

In \cite{HoffmanMahler2004} and \cite{SchuhmacherVoVo2008} the transport idea was used to evaluate 
the performance of multi-object filtering and control algorithms. 
Using a simulated ground truth of a varying number of objects moving through space, 
the online predictions by an algorithm that had only a cluttered version 
of the ground truth available was judged by performance curves over time. 
These curves at any one time were defined as the cost of the optimal transport 
between predicted configuration and ground truth.

\paragraph{Perceived Plant Color}

The EMD was applied for computing color differences between images of different plant species 
by Kendal et \textit{al.} \cite{KendalEtal2013}.
Comparisons showed that these results were largely consistent 
with qualitative assessments by human experts.

\paragraph{Shape Matching}

A fast approximation of the EMD for shape matching 
was introduced by Grauman and Darrell in \cite{GraumanDarrell2004}.
Similar shapes are retrieved by embedding the mimimum weight matching of the contour features of 
a query contour and performing an approximate nearest neighbors search with locality-sensitive hashing.
Ling and Okada proposed a method \cite{LingOkada2007} that reduces the 
computational complexity for computing the EMD between histograms
and they show its usefulness for shape matching and histogram feature matching.
However, the method is restricted to the taxicab metric ($\ell_1$ distance). 

\paragraph{Cell Classification}

Qiu \cite{Qiu2013} considered the two-class problem 
of classifying cells represented by multi-dimensional flow cytometry data
into cells from heathly donors
and cells from patients with acute myeloid leukemia.
The EMD was used by Qiu to compare cell distributions and to derive 
features for classification.

\paragraph{Complex Scene Analysis}

Ricci et \textit{al.} apply the EMD idea for analyzing complex scenes such as frames from videos
which change dynamically and they propose 
to learn a sparse set of prototypes with EMD \cite{RicciZenSebeMesselodi2013}.

\paragraph{Visual Object Tracking}

Zhao et \textit{al.} address the problem of visual object tracking \cite{ZhaoYangTao2010}.
They argue that the EMD is suited for capturing the perceptual differences between images, however, 
its computational complexity is too large for many potential applications.
They propose a differential EMD for tracking which has a reduced computational complexity.

\paragraph{Squared Euclidean Distances and the Interpolation of Shapes and Images}

In the last two decades, numerical schemes were proposed for the special situation
that the ground distance is the square of the Euclidean distance
between origins and destinations.
Aurenhammer et \textit{al.} \cite{AurenhammerHoffmannAronov1998} proposed an algorithm which 
uses power diagrams to transform the transportation problem into 
an unconstrained convex minimization problem. 
Recently, M\'erigot \cite{Merigot2011} improved this algorithm by solving this optimization problem 
via a multiscale approach and applied it to the interpolation of images. 
Further methods for solving transportation problems
with a squared Euclidean ground distance were proposed by 
Benamou and Brenier \cite{BenamouBrenier2000},
by Angenent et \textit{al.} \cite{AngenentHakerTannenbaum2003}, 
by Loeper and Rapetti \cite{LoeperRapetti2005} 
and by Benamou et \textit{al.} \cite{BenamouFroeseOberman2014}.

\paragraph{Assignment Problems}

An important special case of the transportation problem is the assignment problem,
where the numbers of origins and destinations are the same
and the mass at each origin and destination is equal to one.

There exists a multitude of applications in computer science and electrical engineering 
as well as in operations research: e.g. assigning $n$ persons to $n$ jobs, 
or $n$ computational tasks to $n$ nodes in a network. 

For geographical coordinates obtained at different points in time 
for objects like airplanes from radar or satellites, target tracking can be viewed as an 
assignment problem by matching moving targets observed at two points in time. 
However, if more than two points in time are considered simultaneously,
the problem becomes a multi index assignment problem which is a NP-hard problem \cite{SpieksmaWoeginger1996}.

Further potential applications can arise in the area of future public transportation systems:
in case of a prevalence of electric drive vehicles and autonomous driving,
the proposed method can be used to optimally assign cars to recharging locations, 
using for recharging e.g.\ a wireless transmission by electromagnetic induction.

\section*{Acknowledgment}
C. Gottschlich gratefully acknowledges the support of the 
Felix-Bernstein-Institute for Mathematical Statistics in the Biosciences
and the Volkswagen Foundation.


\end{document}